# GPLAC: Generalizing Vision-Based Robotic Skills using Weakly Labeled Images

Avi Singh, Larry Yang, Sergey Levine
University Of California, Berkeley

## Abstract

*We tackle the problem of learning robotic sensorimotor control policies that can generalize to visually diverse and unseen environments. Achieving broad generalization typically requires large datasets, which are difficult to obtain for task-specific interactive processes such as reinforcement learning or learning from demonstration. However, much of the visual diversity in the world can be captured through passively collected datasets of images or videos. In our method, which we refer to as GPLAC (**G**eneralized **P**olicy **L**earning with **A**ttentional **C**lassifier), we use both interaction data and weakly labeled image data to augment the generalization capacity of sensorimotor policies. Our method combines multitask learning on action selection and an auxiliary binary classification objective, together with a convolutional neural network architecture that uses an attentional mechanism to avoid distractors. We show that pairing interaction data from just a single environment with a diverse dataset of weakly labeled data results in greatly improved generalization to unseen environments, and show that this generalization depends on both the auxiliary objective and the attentional architecture that we propose. We demonstrate our results in both simulation and on a real robotic manipulator, and demonstrate substantial improvement over standard convolutional architectures and domain adaptation methods.*

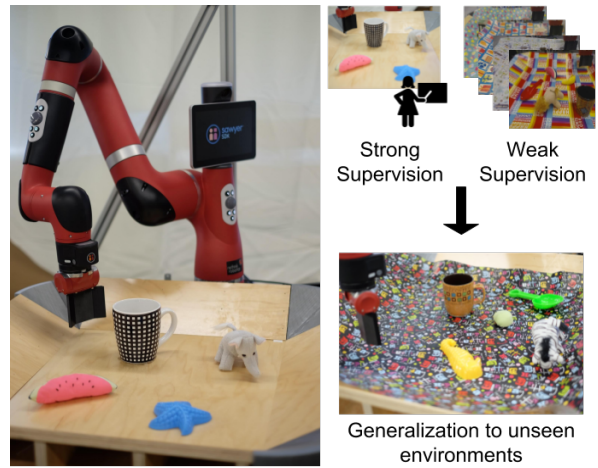

Figure 1. Illustration of our problem setup. A robot (left) must learn a sensorimotor skill, which in this case involves pushing a mug. The robot receives strong supervision for a single mug in a single environment, which here consists of example demonstrations. The robot is also provided with a set of weakly supervised images, which are not collected from a robotic interaction scenario, where the label only signifies whether the image contains a mug. From these two sources of data, the robot must learn a sensorimotor policy that will push new mugs in new environments. See video at: http://people.eecs.berkeley.edu/~avisingh/iccv17/.

## 1. Introduction

We can learn about the world through interaction [47] as well as passive perception. Interactive learning forms a crucial part of the development process of humans and animals, famously demonstrated in an experiment on kitten rearing that involved immobilized and freely moving cats constrained to experience similar visual inputs [18]. However, interaction is not the only way that we can acquire knowledge. We can combine our actively collected experiences with passive observations, such as observations of people and events, pictures, and videos, in order to acquire more generalizable and powerful skills.

Just as with human learning, robotic learning can also use both interactively collected and passively observed data. If we want to learn control policies for robots or other cyberphysical systems, we must provide them with experiences that include perception and actuation, either in the form of autonomously collected experience for reinforcement learning, or demonstration data for imitation learning. And just as with human learning, the amount of interactive data that the robot can practically collect is limited, while the amount of visual observations that can be obtained, for example, using images from the Internet, is orders of magnitude larger in both quantity and diversity. For this reason, the conventional approach to robot perception typically makes use of computer vision systems that are trained using standard supervised datasets, which are then manually interfaced with robotic control and planning modules to perform tasks in the real world. However, this systems-level solution to robotic perception does not make use of the close cou-pling between perception and control that can be exploited by end-to-end methods to improve accuracy, dexterity, and robustness [27, 9, 52, 11, 39]. Can we combine end-to-end robotic learning with the generalization power that can be obtained from large passively collected datasets? We study this question, and propose a technique inspired by multitask learning and spatial attention models that combines the benefits of end-to-end learning from interaction data with the diversity of large, weakly labeled image datasets.

In our proposed problem formulation, illustrated in Figure 1, a sensorimotor policy is trained from two sources of data: interactive data, which consists of example demonstrations or autonomously collected experience from the robot itself, and passive data, which consists only of pictures of objects that are involved in the manipulation task. For example, the interactive dataset might include motor commands and images for a robot pushing a mug onto a coaster, while the passive dataset might consist of a diverse set of pictures of different mugs, along with negative examples of images that do not contain mugs. In essence, the interactive data tells the algorithm *how* to perform a task, while the passive dataset tells it *what* to perform that task on, illustrating the types of objects over which the robot should and should not generalize.

This problem combines both partial supervision and domain shift: the passive data lacks the supervision required to learn the actions, and it is generated from a different distribution, since it does not include the interaction itself. For example, in a robotic manipulation task, the images in the interaction data might include the robot's arm, and might be arranged into video sequences showing coherent movements. The passive data might contain only still images showing the objects without the robot's arm, with a different distribution of positions, poses, and camera viewpoints. While a number of prior methods have proposed to tackle either semi-supervised learning with partial supervision [23, 42, 57, 55] or domain shift [10, 2, 5], tackling both problems at once is very difficult.

The main contribution of this work is to demonstrate that passively collected data can be paired with interaction data to learn visual representations for end-to-end control policies that generalize substantially better to unseen environments. GPLAC uses a spatial attention layer based on the spatial soft argmax operator to ignore irrelevant distractors in the scene, facilitating transfer and generalization in the presence of domain shift. To incorporate the weakly labeled data, the model is simultaneously trained on two tasks: classification of weakly labeled images against negative examples (i.e.,"is there a mug in this image?") and action prediction (i.e., "which motor command should I apply in order to push the mug into the right location?"). We show that, even if our interaction data comes from a single setting (e.g. pushing a single mug), the resulting policy can generalize to new settings, such as new mug appearances and backgrounds. Our experiments are conducted on simulated manipulation tasks and on a real Sawyer robotic manipulator. We demonstrate that both the spatial attention mechanism and the multitask training setup with weakly labeled data are essential to obtain good generalization with minimal interaction data. We further show that our approach outperforms prior semi-supervised methods and domain adaptation techniques, as well as more standard convolutional models that do not use an attentional mechanism.

## 2. Related Work

**Sensorimotor policy learning.** Although standard methods in robotics and autonomous vehicle control typically separate perception and decision making into separate stages [20, 34, 8, 7], the success of deep learning methods has prompted considerable recent interest in learning sensorimotor policies that directly map image pixels to actions [52, 27, 1, 29, 45, 43, 24, 37, 38], following on classic work on end-to-end imitation and reinforcement learning from the 1980s and 90s [40].

In contrast to the standard approach that separates perception and control, these integrated methods allow the perception system to adapt to the needs of the task, simplifying the system design and producing improved performance [28, 39]. Prior work has shown end-to-end learning methods that can acquire complex skills [27] and achieve generalization [1, 39]. However, both separation of perception and control and the end-to-end approach, which combines them into a single policy, have significant shortcomings. The former does not allow the perception system to adapt to the control task, while the latter does not leverage passively collected data. Although finetuning can provide some degree of transfer, we demonstrate in Section 6 that our approach can substantially improve generalization. Prior methods have also sought to overcome the robotic data challenge with brute force, pooling large volumes of data from extensive robotic experience [39, 28] or simulation [19, 16, 33]. While these methods produce effective sensorimotor skills, they are cumbersome to use, and the diversity of collected data is still lower than what can be obtained from entirely passive image datasets.

**Semi-supervised learning.** One way to merge strongly labeled and weakly labeled or unlabeled data is semi-supervised learning, which uses the unlabeled data to construct representations and discover invariances that can allow the supervised learning to achieve better generalization. Models such as variational autoencoders [23, 32, 41], ladder networks [42], stacked what-where autoencoders [57, 55], and generative adversarial networks [44] offer different approaches to combining density estimation or reconstruction objectives with small numbers of labeled images. A difference between our work and semi-supervised learning is that, while semi-supervised learning assumes that the unlabeled data is from the same distribution as the labeled data, this assumption generally does not hold when combining active and passive data in a robotic manipulation setting. The active data involves interaction, almost always contains the robot itself in the image, and is biased toward a small region of the space of images, while the passively collected data is more diverse, might contain differences in viewpoint, and often does not contain a robot in the picture at all. In our experiments in Section 6, we found that CNN-based models without attention perform poorly on our task even with extensive supervision, but we do compare to a semi-supervised method with attention based on deep spatial autoencoders [9], and show that our approach substantially outperforms this prior method.

**Unsupervised domain adaptation.** One way to mitigate the domain shift between actively and passively collected data is domain adaptation, which tries to discover features that are invariant between the two domains and therefore ignore any extraneous factors of variation. Prior methods for domain adaptation have used adversarial training or metrics such as maximum mean discrepancy (MMD) to either learn representations of data that are domain invariant [50, 31, 10, 2, 5], or have learned a transformation function between the two domains [30, 4, 48]. We compare our approach to domain adaptation via gradient reversal [10, 2], and find that GPLAC outperforms this technique in our problem setup. In contrast to standard domain adaptation settings, in our problem setting the factors of variation between the two domains are often relevant to the task. For example, in the mug example in Figure 1, the passive mug images will likely have a different distribution of poses than the active data, and it is precisely the mug position that matters most for the task. Therefore, simply enforcing domain invariance runs counter to the goal of acquiring useful and generalizable representations from the weakly labeled data.

**Multitask learning** Multitask learning is often used to tackle related problems in computer vision, as it helps transfer information across different datasets to achieve superior performance [12]. A standard approach is to train convolutional neural networks with some layers shared across tasks, and some layers specific to the individual tasks [13, 51, 56, 17]. . While our approach is most similar to multitask learning, we are not focusing on the multitask learning problem itself: the goal of our system is not to learn *both* a policy and a binary classifier, but to enable a policy to generalize better using the additional signal obtained from binary classifier training on the weakly labeled data.

**Visual attention** Our spatial attention mechanism performs an approximate localization operation, making it well suited for transferring features from the weakly labeled data to the policy learning task. Visual attention has been used in a number of computer vision settings, such as image captioning [53], visual question answering [54], and image generation [15]. Attention mechanisms often use a recurrent network which decides where the network should look at each time step to solve the task at hand. However, while most previous work produces attention maps, we go a step further and take a soft arg-max of the attention map to extract only the spatial information. This results in the spatial location of the most salient points in the image being passed to the subsequent layers.

## 3. Problem Formulation

The goal of our method is to enable learning of generalizable sensorimotor skills from a combination of interactive experience and passively collected data. We can define this problem as one of policy learning: the goal is to acquire a policy model $\pi_\theta : \mathcal{O} \to \mathcal{A}$ that maps from the agent's observations $o \in \mathcal{O}$ to actions $a \in \mathcal{A}$. This policy might involve, for example, having a robotic agent process images from its camera, localize objects of interest, and output motor commands to its actuators to perform a desired task. Such policies would typically be trained with reinforcement learning or learning from demonstration. In either case, we can define a generic optimization for the policy parameters as

$$\theta^\star \leftarrow \arg\min_\theta \mathcal{L}_{\text{task}}(\pi_\theta),$$

where $\mathcal{L}_{\text{task}}(\pi_\theta)$ could be the negative reward in reinforcement learning, or the difference between the actions of $\pi_\theta$ and a set of expert demonstrations. We will use the learning from demonstration setup for the derivations in this paper, but the framework can easily be extended to reinforcement learning simply by using a different loss $\mathcal{L}_{\text{task}}(\pi_\theta)$. For learning from demonstration, we can define $\mathcal{L}_{\text{task}}(\pi_\theta)$ as

$$\mathcal{L}_{\text{task}}(\pi_\theta) = \frac{1}{N}\sum_{i=1}^N d(\mathbf{a}_i, \pi_\theta(\mathbf{o}_i)),$$

where $d(\cdot, \cdot)$ is some distance metric, such as Euclidean distance, between the demonstrated actions $\mathbf{a}_i$ and the action taken by the policy at the corresponding observation $\mathbf{o}_i$. The goal of learning the sensorimotor skill $\pi_\theta$ is not just to achieve mastery in a particular narrow range of settings, but to generalize to new situations that were not seen during training. To that end, we can define a set of labeled environments $E_l$ and a set of unseen environments $E_u$. All of the data used for $\mathcal{L}_{\text{task}}(\pi_\theta)$ comes from $E_l$, but the policy must also achieve good performance when executed on environments in $E_u$. Variation between environments might include the appearances and identities of objects, background, and lighting, though we assume there are minimal physical discrepancies. For example, as shown in Figure 5, different environments for an object pushing task might have different distractor objects and differences in the appearance of the main object of interest (in this case, a coffee cup that must be pushed onto a coaster). If $E_l$ itself is large and highly varied, a policy trained on $E_l$ will be able to achieve generalization to the new environments in $E_u$. However, collecting active interaction data or example demonstrations in a huge variety of environments can be expensive and time consuming. It would be much more efficient to learn highly generalizable sensorimotor policies by combining interactive data from $E_l$ with passively collected data.

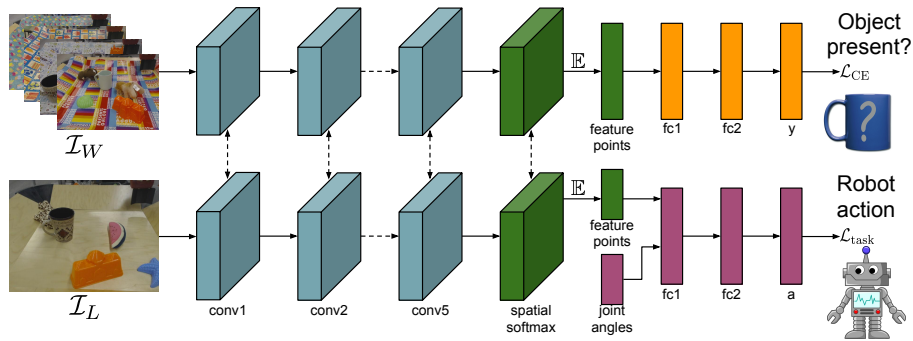

Figure 2. GPLAC architecture. We train two convolutional neural networks: the policy $\pi_\theta$, and the binary classifier $c$. The two networks share their convolutional layers (shown in blue), and have separate fully connected layers (shown in orange and magenta). Our spatial attention layer (shown in green) lies between the convolutional layer and the fully connected layer, and forms a major information bottleneck. For predicting an action, the robot's state information (joint angles, velocities, end effector position) are also passed into the network $\pi_\theta$. We train the policy $\pi_\theta$ using our expert demonstrations and the loss $\mathcal{L}_{\text{task}}$, while the classifier is trained with the weakly labeled images and their binary class labels.

To incorporate passively collected data into this learning process, we will assume that we also have access to a set of weakly labeled environments $E_w$. For these weakly labeled environments, we can obtain observations $\mathbf{o}_j$, but not the corresponding actions $\mathbf{a}_j$. We therefore cannot evaluate either reinforcement learning or imitation learning loss functions on the samples from $E_w$. We will, however, assume that the samples from $E_w$ are weakly labeled with a label $y$ that indicates simply whether or not the corresponding observation $\mathbf{o}$ does or does not contain the object of interest, such as the coffee cup in Figure 1. Intuitively, providing the system with weakly labeled data of this type tells it which types of objects it should generalize to, while the labeled data from $E_l$ tells the policy how to perform the task.

## 4. Learning from Weakly Labeled Data

Unlabeled or weakly labeled data can be incorporated into the learning process using semi-supervised learning methods, as discussed in Section 2. However, in the presence of significant domain shift between the labeled and weakly labeled domains, semi-supervised methods that assume the same input distribution tend to perform poorly. Domain adaptation methods, also discussed in Section 2, can alleviate this problem, but introduce the problematic assumption that factors of variation that correlate with domain identity are irrelevant to the task (domain invariance). This is not typically the case for sensorimotor skills, where variables like the positioning of objects might vary systematically between the two domains and are highly relevant. We instead adopt a simple multitask learning approach to incorporating weakly labeled data, and we observe in Section 6 that this simple approach is highly effective at transferring generalizable features and appearance invariance from the weakly labeled data. The intuition is that the active labeled data from $E_l$ tells the algorithm *how* a particular task should be performed, while the weakly labeled data from $E_w$ tells it which objects it should be performed on. By observing the positive and negative examples in $E_w$, the method can figure out over which features it should or should not generalize. The training objective is simply a linear combination of the task loss $\mathcal{L}_{\text{task}}$ on $E_l$ for action prediction and $\mathcal{L}_{\text{task}}$ on $E_w$ for binary classification, with a weight $\lambda$ on the secondary classification task

$$\mathcal{L} = \mathcal{L}_{\text{task}}(\pi_\theta) + \lambda \mathcal{L}_{\text{CE}}(c_\theta),$$

where $c_\theta(\mathbf{o})$ is a binary classifier that shares some parameters with $\pi_\theta(\mathbf{o})$, and $\mathcal{L}_{\text{CE}}$ is a binary cross-entropy classification loss The model that we use to represent $\pi_\theta$ and $c_\theta$ is a convolutional neural network with a spatial attention mechanism, which is described in detail in the next section. The network has shared convolutional weights between $\pi_\theta(\mathbf{o})$ and $c_\theta(\mathbf{o})$, with separate weights for the fully connected layers of the policy and classifier, as illustrated in Figure 2. A conventional CNN trained with this multitask objective does not attain good performance, as shown in Section 6. The attentional mechanism described in the next section enables generalization by forcing the network to share representations between the classification and policy learning tasks.

## 5. Generalization Through Spatial Attention

Conventionally, end-to-end policies that directly process image pixels have used convolutional neural networks similar to those employed for image recognition [25, 26, 3, 6]. However, when labeled data is scarce, using more parameter efficient architectures can be advantageous. In this work, we build on a spatial attention mechanism proposed in prior work based on a special soft-argmax [27, 9, 14]. While prior work used this mechanism for learning policies in a parameter efficient way for a single environment [27], we focus specifically on how it can be adapted to facilitate generalization with weakly labeled data. We show that this mechanism greatly improves the ability of the model to generalize under variation in the appearance of the background, foreground object, and distractors, enabling it to perform the learned skill in new, visually distinct environments.

The architecture is shown in Figure 2. The spatial attention layer forms the interface between the convolutional and fully connected layers. It consists of a "soft" approximation to the spatial arg-max operation, computing a soft estimation of the image-space point corresponding to the maximal activation of each channel in the last convolutional layer. Letting $h_{cij}$ denote the activation in the $c^{\text{th}}$ channel of the last convolutional layer at the image-space position $(i,j)$, the spatial attention layer first computes a spatial softmax according to

$$z_{cij} = \frac{e^{h_{cij}}}{\sum_{i',j'} e^{h_{ci'j'}}},$$

and then computes the expected feature location corresponding to the channel $c$ as

$$\mathbf{f}_c = \left( \sum_i i * z_{cij}, \sum_j j * z_{cij} \right)$$

Thus, the output of this attentional mechanism is a vector of activations with length $2c$ – two coordinates for each channel. Examples of points produced by this attention layer are shown in Figure 5. This attention mechanism creates a strong information bottleneck between the convolutional and fully connected layers. The lateral inhibition imposed by the spatial softmax also greatly reduces the influence of distractors, since non-maximal activations in each convolutional layer are suppressed. This means that, if each channel of the last convolutional map is sensitive to a particular feature in the image, distractors that activate that feature less than the object of interest will have minimal impact. This allows the network to narrowly "attend" just to the parts of the image that are most relevant. As we will show in Section 6, this allows the attentional architecture to substantially outperform more standard convolutional models.

This architecture also helps to facilitate transfer, since the binary classifier in Section 4 must use the same features as the policy to detect whether the object of interest is present in the weakly labeled data. This implicitly forces the feature points to act as detectors for the object of interest, ignoring other distractors and learning invariance to background variation. The illustration in Figure 5 confirms this: note that many of the feature points are on the object of interest (the mug), which means that the mug has the highest activations in the last convolutional layer. Furthermore, in order to be able to reliably classify the weakly labeled images that do not contain the object of interest, the network must commit multiple channels to attend to the object of interest, since the subsequent fully connected layers only observe the positions of the feature points, not their content. The network must therefore ensure that images that contain the object produce a consistent, structured pattern of feature point positions that can be easily distinguished from the chaotic patterns produced by the negatively labeled images. While image classification models learned on large scale datasets can also localize objects without any explicit supervision [35], GPLAC can learn to attend to objects of interest with relatively modest amounts of weakly labeled data, and produce a representation that is also useful for learning the corresponding sensorimotor skill.

## 6. Experimental Evaluation

We evaluate GPLAC on two tasks in simulation (with rendered images), and one task on a real robot. The experiments aim to determine whether GPLAC can learn policies that generalize to new environments with active data from just a single environment, compare it to prior semi-supervised learning and domain adaptation techniques, and compare our spatial attention model to standard CNNs. The active task in our evaluation corresponds to learning from demonstration, where the labeled environments include ground truth action labels obtained from an optimal policy (in simulation) or from a user demonstration (on the real robot). However, an extension of this method to reinforcement learning would be straightforward, particularly with reinforcement learning methods that already use supervised learning in the inner loop, such as guided policy search [27] or reward-weighted regression [36]. Please refer to the appendix for details on the dataset, network architecture and learning procedure.

### 6.1. Simulated Experiments

Our simulated experiments use the MuJoCo physics simulator [49], with a realistic rendering frontend that provides for a wide variety of models and textures for objects that can be interacted with by a simulated robotic manipulator. We collect expert trajectories from the reinforcement learning algorithm TRPO [45], that uses the positions of objects in the world as input. After we have collected expert trajectories, we drop the full state information, requiring the senso-rimotor policy trained from these demonstrations to use the raw image pixels to perform the task. The policy is provided with the robot's joint angles and velocities, but all information about the objects in the scene must be inferred from the images. We perform end-to-end learning in this setup, mapping images from the simulated camera to the torques applied at the joints of the robot arm. We use 400 expert trajectories from one labeled environment (100 timesteps each), and weakly labeled data from 40 different environments (one of which is also the labeled environment), with 1000 "positive" images in each environment that contain the object of interest, and 1000 "negative" examples that do not. The weakly labeled images have considerable domain shift with regard to the active demonstration data: they do not contain the robot in the scene, have a different distribution of object locations, and contain small variations in the viewpoint. We use a set of unseen environments as a validation set, and 40 unseen environments as a test.

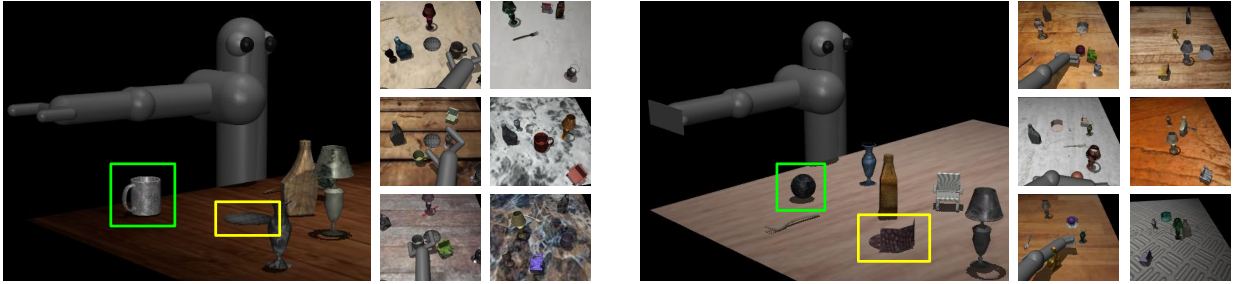

Figure 3. Illustration of the simulated tasks used in our experiments: the pushing task (left) and the striker task (right). The large image shows a third-person view of the robot, while the smaller images show different test environments from the robot's point of view, with the weakly labeled images shown in the right column. Note the variety of backgrounds and object appearances, and the lack of robot arm in the weakly labeled dataset.

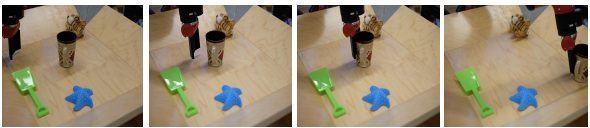

Figure 4. Illustration of our real robot pushing task. The goal is in the corner of the arena, shown in the far right image. The robot first approaches the mug based on the arm pose given as output by the neural network, and then pushes the mug in a straight line. The position of the mug is varied in every trial.

Our simulated evaluation consists of two tasks, shown in Figure 3, both of which require the learned sensorimotor policy to control a 7 DoF arm by applying torques at the joints at 20 Hz. In the first task, the arm must use camera images to push a mug onto a round coaster. The initial position of the mug differs from trial to trial, and the appearance of the mug, background, and distractor objects differs between environments. The policy must determine the position of the mug and guide the end-effector to it in order to perform the task. Note that this task requires more than simply predicting the mug position: raw torque control on a 7 DoF robotic arm requires delicate coordination of the joints to perform the pushing motion, and the policy must learn on its own that it must attend to the mug position in order to successfully emulate the demonstrated behaviors.

The second task requires using the arm to strike a ball and roll it into a goal cup. The goal changes position on each trial, and the appearance of the ball, goal, background, and distractors changes between environments. The policy must learn that it should attend to the goal, servo the arm toward the ball, and learn how to impart the right impulse to roll it into the goal. We compare to a number of baselines, prior methods, and ablated variants of our approach:

- **ATT-1.** Attentional model trained on labeled data from one environment. This is a lower bound on performance without any transfer from weakly labeled data.

- **ATT-40.** Attentional model trained on labeled data from 40 environments. This is an upper bound – the total amount of data is the same as the weakly labeled case, but labels are available for all environments.

- **GPLAC.** Our method, using the attentional model trained on one labeled environment and using the classifier to add weakly labeled data from 40 environments. This is the standard version of our method.

- **GPLAC-DA.** Our method combined with domain adaptation based on [2, 10]. This variant is meant to evaluate whether adding domain adaptation has any benefit beyond our multitask classifier.

- **ATT-DA.** Our attentional model trained with domain adaptation [2, 10] for the weakly labeled data, without the binary classifier. This is a strong baseline that combines our attentional mechanism with domain adaptation.

- **ATT-FT.** Finetuning baseline: attentional model is first trained on for binary classification on the weakly labeled data, and then finetuned on the labeled environment. This is representative of prior finetuning methods.

- **ATT-AE.** Attentional model trained as a spatial autoencoder, following the work of Finn et al. [9].

| Setting | 1 labeled env. | | 40 labeled env. | | 1 labeled env. + unlabeled data from 40 env | | | | | | |
|---|---|---|---|---|---|---|---|---|---|---|---|
| Model | ATT-1 | CNN-1 | ATT-40 | CNN-40 | CNN-C | ATT-AE | ATT-PO | ATT-FT | ATT-DA | GPLAC-DA (our) | GPLAC (our) |
| Pusher | 37.50 | 40.25 | 90.25 | 75.50 | 2.25 | 42.50 | 55.50 | 52.75 | 69.50 | 66.50 | **79.75** |
| Striker | 35.50 | 39.75 | 69.50 | 52.50 | 21.50 | 35.75 | 39.50 | 37.75 | 40.50 | **61.00** | 59.25 |

Table 1. Simulation results. The two left columns show the performance without any weakly labeled data, while the columns for 40 labeled environments show the upper bound performance with 40 labeled environments. GPLAC achieves the best results without domain adaptation on pusher and the best results with adaptation on striker (GPLAC-DA), though GPLAC is very close. This suggests that domain adaptation has minimal benefit on these tasks on top of our method. The prior semi-supervised learning (ATT-AE) and domain adaptation (ATT-DA) methods achieve lower improvement, even using our attentional mechanism. The standard CNN methods achieve poor results on all tasks, even when provided with more supervised data than our approach (CNN-40).

- **ATT-PO.** Attentional model is first trained to regress pose of the objects, following the work of Levine et al. [27]. The model is then finetuned on the labeled environment. Note that neither GPLAC nor any of the other baselines have access to pose information.

- **CNN-1.** Standard convolutional neural network model (without attention) trained on labeled data from just a single environment.

- **CNN-40.** Standard convolutional neural network model (without attention) trained on labeled data from all 40 environments. This is an upper bound on the performance of a standard CNN model.

- **CNN-C.** Standard convolutional neural network model (without attention) trained with a single labeled environment and weakly labeled data from 40 environments, using the binary classifier to incorporate the weakly labeled data. This is an ablated version of our method that does not use the attentional mechanism.

The results in Table 1 show that GPLAC achieves the best generalization to unseen environments on both tasks, within about 10% of the best possible performance obtained when training on 40 labeled environments. Surprisingly, our approach with just one labeled environment outperforms a conventional CNN model even when that model is trained with labels on all 40 environments, suggesting that the strong lateral inhibition and distractor rejection of the attentional model greatly improves its ability to generalize to unseen settings. The non-attentional CNN with our classifier (CNN-C) performs much worse on both tasks, likely due to the CNN allocating different units to handle the action prediction and classification tasks, instead of benefitting from their shared structure. This sharing is enforced in the attentional model because of the narrow bottleneck of the feature points. Due to the poor performance of the CNN, we expect that prior semi-supervised methods based on standard, non-attentional CNNs, such as ladder networks [42] or variational autoencoders [21, 23] are unlikely to perform well. We compare instead to a semi-supervised method based on the spatial autoencoder, which also uses spatial attention [9]. However, this method also doesn't perform as well as our approach. The most com-petitive baseline is the attentional model with domain adaptation (ATT-DA) [2, 10], which performs about 10% worse than GPLAC on both tasks. Interestingly, adding domain adaptation to our method (GPLAC-DA) produces very little improvement, and on only one of the tasks. This suggests that most of the benefit of the weakly labeled data is already obtained from the binary classifier, and there is no need for additional domain adaptation.

| Model | ATT-1 | ATT-DA | GPLAC |
|---|---|---|---|
| Success Rate | 33.22 | 45.45 | **56.36** |

Table 2. Real robot results for ATT-1, ATT-DA (domain adaptation with our attentional architecture), and GPLAC (our method). GPLAC achieves over 20% improvement in generalization to new mugs and backgrounds from utilizing the weakly labeled data.

| Task | 80000 | 8000 | 800 | 80 |
|---|---|---|---|---|
| Pusher | 79.75 | 79.50 | 78.50 | 56.00 |
| Striker | 59.25 | 56.75 | 53.00 | 33.50 |

Table 3. Simulation results with reduced weakly labeled data. All numbers are success rates. Performance remains high with upto 800 images, and drops sharply on further reducing the number of images.

**How many images do we need?** We conducted further experiments to find out how many images are needed to obtain effective generalization. The results in Table 6.1 show that the performance remains high with upto 800 weakly labeled images images (which amounts to 20 images per environment, in addition to the 40K images obtained from expert trajectories from a single environment), and we believe this is due to the parameter-efficient nature of the spatial attention mechanism.

### 6.2. Real Robot Experiments

Our real-world experiments are performed on a Rethink Robotics Sawyer robot. The task, shown in Figure 4, is to push a mug from a random position in the workspace to a goal location. The action is specified in terms of the starting point for a pushing motion, from which the robot moves the arm in a straight line towards the goal. We collected 306 example trajectories for a single mug in a single environment. We also collected unlabeled data with 17 different mugs and four different backgrounds, producing 50 scenes that contain a mug and 50 negative examples, with multiple images from each scene that have the mugs in different locations. The result is 1700 images for the weakly labeled classification task. Testing is performed on *unseen* mugs, for which neither expert demonstrations nor weakly labeled images were available to the method. We evaluate GPLAC on 11 unseen mugs, with unseen distractors and backgrounds, and 10 trials per mug. The results in Table 2 show that our method achieves good performance on this task, comparable to the success rates attained in the simulated experiments. Since the real-world robot experiments are substantially more time consuming, we only compared against the strongest baseline from the simulated experiments (ATT-DA). The performance of this baseline fell halfway between our method and the attentional network trained without any weakly labeled data. This confirms that our approach produces better generalization than standard domain adaptation, and can greatly improve the generalization abilities of a sensorimotor policy for a real-world robotic manipulation task.

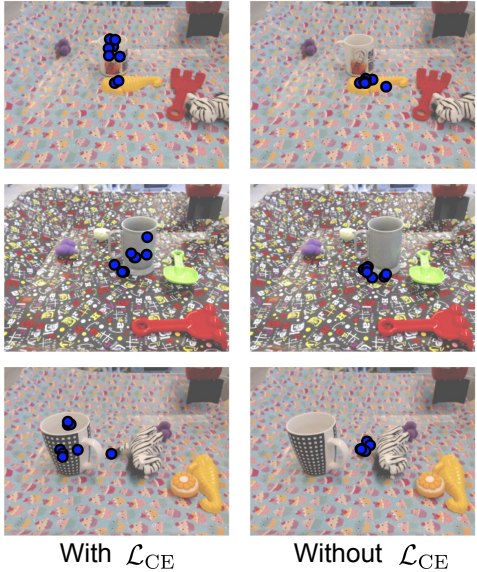

Figure 5. Illustration of feature points $\mathbf{f}_c$ for various mug images with GPLAC (left) and ATT-1 (right). Introducing the classification task (left) causes the feature points (blue circles) to cluster on the object of interest (the mug), while training only on the single labeled environment produces feature points that are easily distracted by other objects (top) and complex background patterns (middle and bottom).

To illustrate the behavior of the spatial attention model, we also visualize the feature points of GPLAC, as well as ATT-1, in Figure 5, for new mugs not seen during training. The feature points for GPLAC are almost entirely clustered on the mug, suggesting that the model was able to successfully learn to localize mugs in general, rather than overfit-ting to the single mug in the labeled data.

## 7. Discussion and Future Work

We proposed a method for training sensorimotor skills that combines active interaction data with passive image datasets to improve the policy's ability to generalize to unseen environments. Our approach relies on two components: an attentional CNN architecture that focuses on objects of interest while ignoring distractors, and a multitask learning setup that trains this network simultaneously on actions and an auxiliary binary classification task on the passively collected images. This auxiliary task amounts to asking the network to predict which images contain the object of interest for the task, and which do not. Intuitively, the active data tells the learner *how* the task should be performed, while the passive data tells it which object it should be performed on. Our results, both in simulation and on a real robotic manipulator, show that our approach achieves large improvements in generalization to unseen objects and backgrounds, and that both the attentional model and the auxiliary classification task are essential to obtain this improvement. In comparisons, our approach outperforms domain adaptation, semi-supervised learning with autoencoders, and non-attentional CNN models.

While GPLAC achieves good performance, it has a few limitations. First, the attentional mechanism is a strong information bottleneck, and while it is sufficient for tasks that are primarily spatial (e.g., localizing a mug in order to push it across a table), it remains to be seen how well this architecture will translate to tasks that involve complex reasoning about object semantics. Second, our multitask training setup still requires weak supervision in the passive data. While this supervision may be easy to obtain in many cases, for example by searching for images of the object of interest on the internet, it still constitutes an additional requirement beyond standard semi-supervised methods. Finally, our evaluation presents results for imitation learning, and we leave it to future work to demonstrate the approach with reinforcement learning from scratch. Such an extension is likely to be straightforward, and would enable substantial improvements in the ability of autonomously learned sensorimotor policies to generalize to new settings.

**Acknowlegements**  This research was made possible by an ONR Young Investigator Program award and support from NVIDIA and Google. We thank Roberto Calandra for assistance with robot experiments and providing feedback on the manuscript. We thank Abhishek Gupta for help with the MuJoCo simulator. We thank Erin Grant, Chelsea Finn, Deepak Pathak and Kate Rakelly for providing their comments on the manuscript. We also thank Shubham Tulsiani and Taesung Park for helpful discussions.

# 8. Appendix

## 8.1. Project website

For more materials associated with this project, including a demonstration video, please visit: http://people.eecs.berkeley.edu/~avisingh/iccv17/.

## 8.2. Dataset

We collected several datasets for our experiments, which we present in this section. The data used for both real robot experiments and simulation experiments is presented here.

### 8.2.1 Real robot

Our robot experiments were carried out using 306 expert demonstrations. The training environment is shown in Figure 6. We used 1700 images as weakly labeled data, a small sample from which is available in Figure 7. We evaluated our policies in unseen environments, which can be seen in Figure 8.

### 8.2.2 Simulation

Our simulation datasets consists of 400 expert trajectories from a single environment, where each trajectory has 100 timesteps. This results in 40,000 images for each task. For the weakly data, we have 2,000 images from each of the 40 environments, which results in a total of 80,000 images.

**Pushing** For the pushing task, the images from training and test environments can be seen in Figure 9. The unlabeled images for the pushing task can be seen in Figure 10. This pushing task has now been merged into the OpenAI Gym repository https://github.com/openai/gym/blob/master/gym/envs/mujoco/pusher.py.

**Striking** For the striking task, the images from training and test environments can be seen in Figure 11. The unlabeled images for the striking task can be seen in Figure 12. This pushing task has now been merged into the OpenAI Gym repository https://github.com/openai/gym/blob/master/gym/envs/mujoco/striker.py.

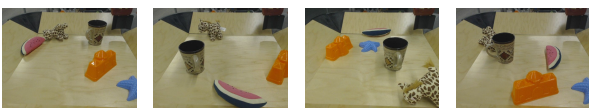

Figure 6. Images from the training environment for the real robot. Note that all the images in this set have the same object of interest (i.e. same mug), and same set of distractors. We collect 306 of such image/action pairs for our real robot experiments.

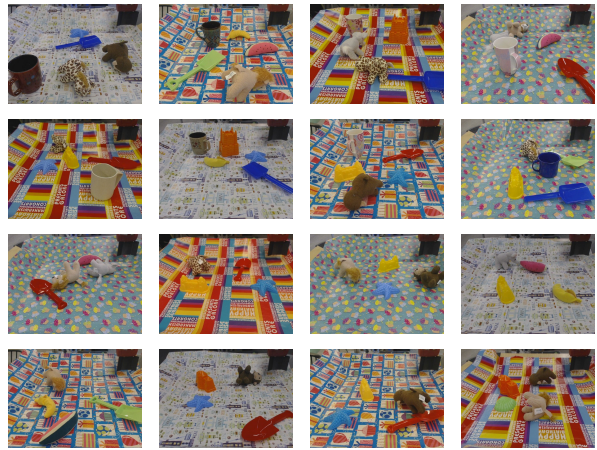

Figure 7. A sampling of the weakly labeled images for the real robot experiments. The top two rows of images contain the mug, while the bottom two do not contain the mug.

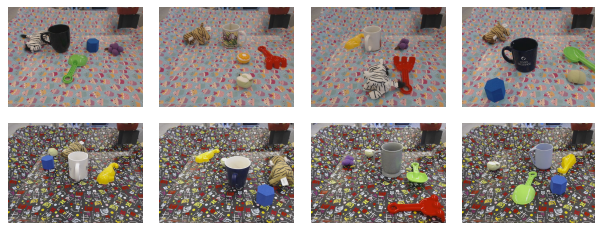

Figure 8. A sample of the test environments for the real robot experiments. Note that the mug being manipulated, the background, and the distractor objects were all unseen during training - in both the expert demonstration and weakly labeled images.

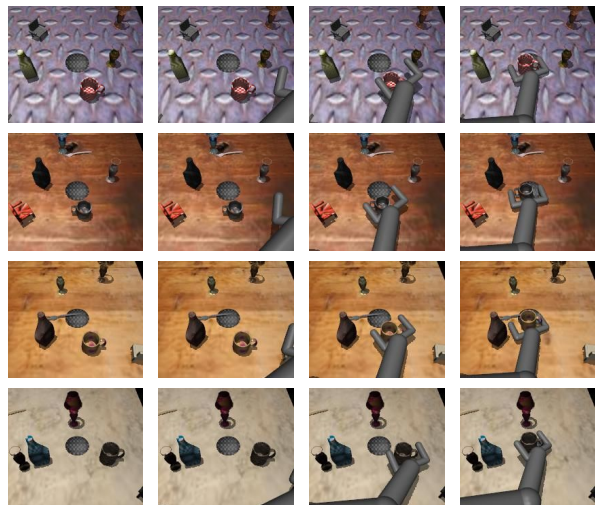

Figure 9. Some example trajectories of the pushing task for from the training and testing environments. The images shown in the first row are from the training environment, while the three rows below that are examples of the test environments - with different textures, lighting conditions, and positioning of the objects.

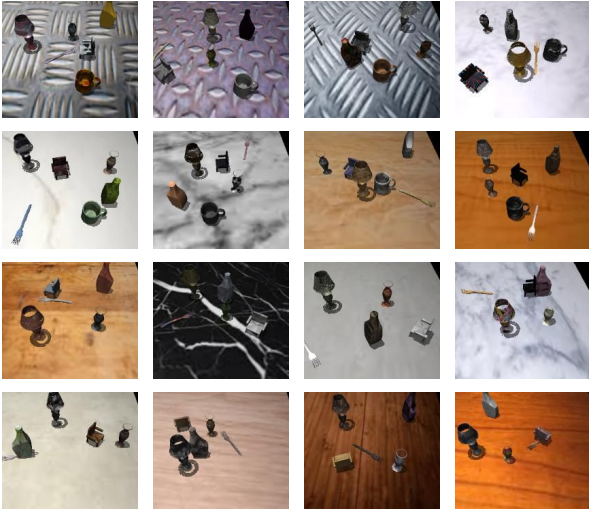

Figure 10. A sampling of the weakly labeled images for the simulation pushing task. The top two rows of images contain the object of interest, while the bottom two do not. Note that none of these images contain the robot arm, and have small viewpoint variations as compared to the expert demonstration images.

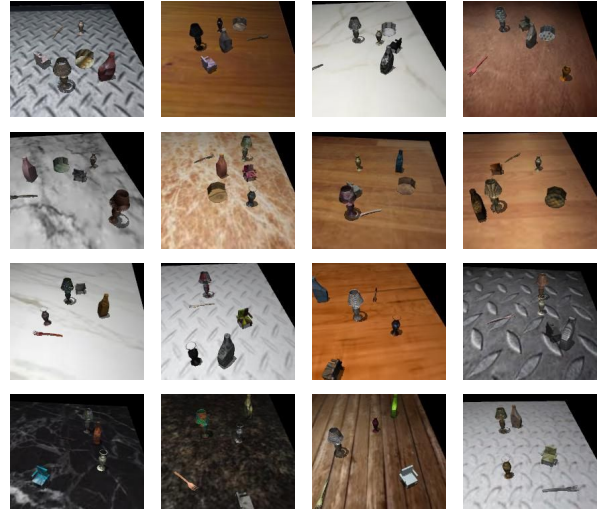

Figure 12. A sampling of the weakly labeled images for the simulation striking task. The top two rows of images contain the object of interest, while the bottom two do not. Note that none of these images contain the robot arm, and have small viewpoint variations as compared to the expert demonstration images.

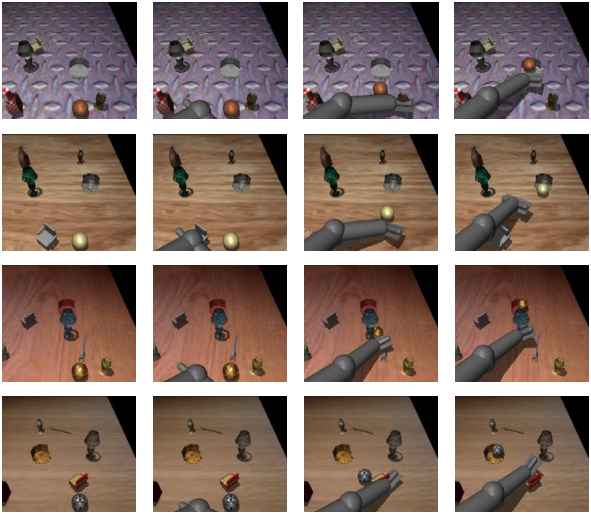

Figure 11. Some example trajectories of the striking task for from the training and testing environments. The images shown in the first row are from the training environment, while the three rows below that are examples of the test environments - with different textures, lighting conditions, and positioning of the objects.

### 8.3. Training Details

All of our models with attention have the same architecture: five convolutional layers with 3x3 filters, and the number of filters are 64, 64, 32, 32, and 16. The stride is equal to 2 for the first conv layer, and 1 for all the subsequent conv layers. The conventional convolutional network instead has the same number of filters and layers as the attention model, a stride of 4 in the first layer, and strides of 1 in subsequent convolutional layers, in order to preserve the spatial information that is essential for the task. The first layer of all networks (with and without our attention mechanism) is initialized from the VGG-16 network [46]. The fully connected layers all have 400 units. We use dropout on the output of the spatial attention layer, in order to force some degree of redundancy into the feature points, which increases the robustness of the model. We also use batch normalization between the convolutional layers. Training is done with Adam [22] and a learning rate of 3e-4, and all models are trained for 50K training steps. Input images are 200x175 for the simulation experiments, and 320x240 for the real robot experiments. For the first 5K steps, we only optimize for $\mathcal{L}_{\text{task}}$; we optimize for the complete objective in all subsequent steps.